\documentclass[letterpaper]{article} 
\usepackage[test]{aaai23}  
\usepackage{times}  
\usepackage{helvet}  
\usepackage{courier}  
\usepackage[hyphens]{url}  
\usepackage{graphicx} 
\usepackage{natbib}  
\usepackage{caption} 
\usepackage{algorithm}
\usepackage{algorithmic}
\usepackage{xcolor}

\newcommand{\xmark}{\ding{55}}%
\usepackage{newfloat}
\usepackage{listings}
\DeclareCaptionStyle{ruled}{labelfont=normalfont,labelsep=colon,strut=off} 
\floatstyle{ruled}
\newfloat{listing}{tb}{lst}{}
\floatname{listing}{Listing}
\usepackage{subcaption}
\usepackage{adjustbox}
\usepackage{multirow}
\usepackage{xr}
\usepackage{threeparttable}
\usepackage{bbding}
\usepackage{times}
\usepackage{epsfig}
\usepackage{graphicx}
\usepackage{amsmath}

\newcommand{\eg}{\textit{e}.\textit{g}.}
\newcommand{\ie}{\textit{i}.\textit{e}.}

\usepackage{amssymb}
\usepackage{pifont}
\usepackage{booktabs}
\usepackage[pagebackref,breaklinks,colorlinks]{hyperref}
\usepackage{graphicx, amsmath, amssymb, caption, subcaption, multirow, overpic, textpos}
\usepackage[british, english, american]{babel}

\newlength\savewidth\newcommand\shline{\noalign{\global\savewidth\arrayrulewidth
  \global\arrayrulewidth 1pt}\hline\noalign{\global\arrayrulewidth\savewidth}}
\newcommand{\tablestyle}[2]{\setlength{\tabcolsep}{#1}\renewcommand{\arraystretch}{#2}\centering\footnotesize}
\renewcommand{\paragraph}[1]{\vspace{1.25mm}\noindent\textbf{#1}}

\usepackage{multicol}
\newcommand{\app}{\raise.17ex\hbox{$\scriptstyle\sim$}}

\definecolor{deemph}{gray}{0.6}

\definecolor{baselinecolor}{gray}{.9}

\title{VLMAE: Vision-Language Masked Autoencoder}

\author{
    Sunan He\textsuperscript{\rm 1,2}\equalcontrib\thanks{This work was done during an internship at Tencent.},
    Taian Guo\textsuperscript{\rm 1}\equalcontrib, 
    Tao Dai\textsuperscript{\rm 3}\corres,
    Ruizhi Qiao\textsuperscript{\rm 1}\corres, 
    Chen Wu\textsuperscript{\rm 1},
    Xiujun Shu\textsuperscript{\rm 1},
    Bo Ren\textsuperscript{\rm 1}
}
\affiliations{
    \textsuperscript{\rm 1} YouTu Lab, Tencent
    \textsuperscript{\rm 2} Tsinghua Shenzhen International Graduate School, Tsinghua University 
     \\
    \textsuperscript{\rm 3} College of Computer Science and Software Engineering, Shenzhen University\\
   \{sunanhe, taianguo, ruizhiqiao, chewu, xiujunshu, timren\}@tencent.com, daitao.edu@gmail.com
}

\begin{document}

\maketitle

\begin{abstract}

Image and language modeling is of crucial importance for vision-language pre-training (VLP), which aims to learn multi-modal representations from large-scale paired image-text data.
However, we observe that most existing VLP methods focus on modeling the interactions between image and text features while neglecting the \textit{information disparity} between image and text, thus suffering from \textit{focal bias}.
To address this problem, we propose a vision-language masked autoencoder framework (VLMAE).
VLMAE employs visual generative learning, facilitating the model to acquire fine-grained and unbiased features.
Unlike the previous works, VLMAE pays attention to almost all critical patches in an image, providing more comprehensive understanding.
Extensive experiments demonstrate that VLMAE achieves better performance in various vision-language downstream tasks, including visual question answering, image-text retrieval and visual grounding, even with up to 20\% pre-training speedup.

\end{abstract}

\begin{figure}[ht]
\centering
\includegraphics[width=0.9\linewidth, page=6]{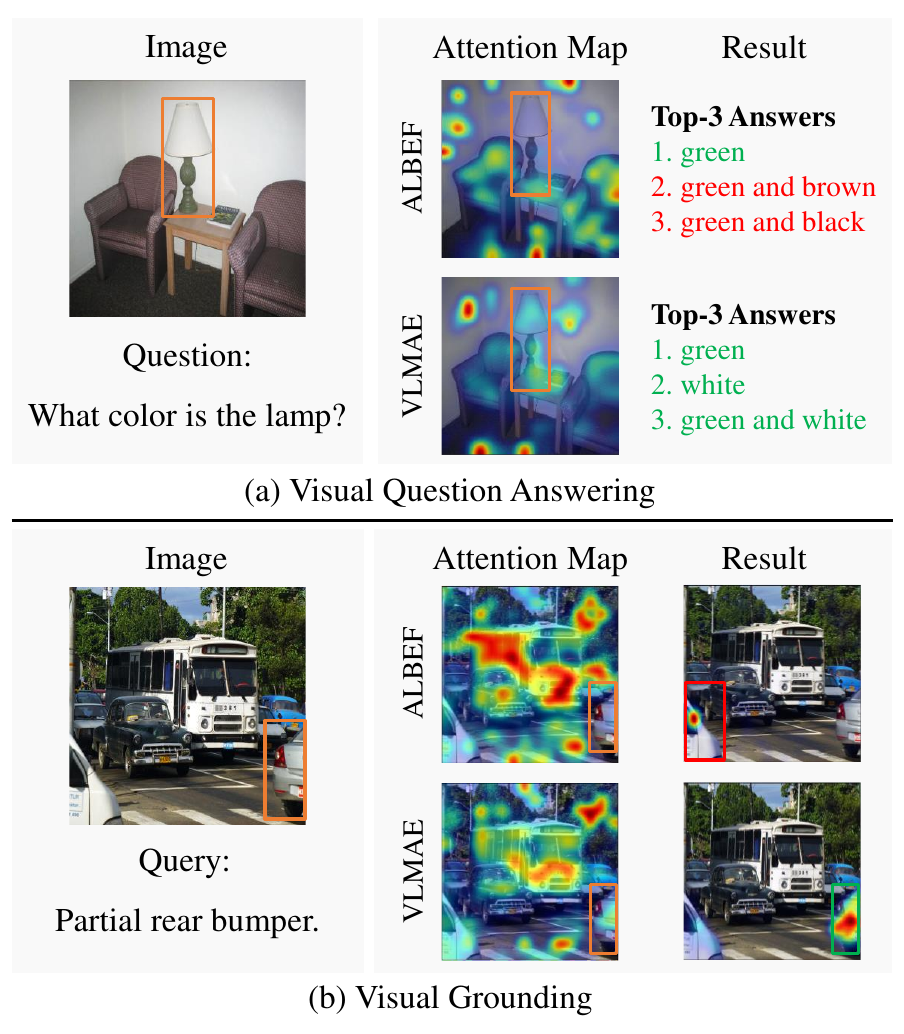}
\caption{
Comparison of ALBEF and our VLMAE in downstream tasks.
ALBEF may focus on specific objects from images, while ignoring other critical objects(\eg, lamp in the top row figure), and thus make wrong predictions (in red). By contrast, our VLMAE produces better results (in green) with a universal focus over the whole image.
} 
\label{fig:teaser}
\end{figure}
\section{Introduction}
In recent years, with the prevalence of self-supervised learning and transformer-based models in both natural language processing and computer vision~\cite{devlin2018bert, dosovitskiy2020image}, we have witnessed rapid development in vision-language representation learning.
In multi-modal tasks, such as visual question answering and image-text retrieval, where the model needs to comprehend and aggregate visual-textual information, vision-language pre-training (VLP) plays a fundamental role.
\par
Early VLP methods~\cite{lu2019vilbert, li2020unicoder, chen2020uniter} rely on a well-trained but heavy object detector on the input image for visual proposal feature extraction, followed by a cross-model encoder for visual-textual interaction.
To eliminate the heavy object detector, some recent methods~\cite{huang2021seeing, kim2021vilt, wang2021vlmo} resort to vision dictionary with a lightweight visual embedder or modality experts networks for better trade-off between efficiency and performance of downstream tasks.
Very recently, ALBEF~\cite{li2021align} introduces contrastive learning to align visual-textual features, facilitating the aggregation of multi-modal information. 
TCL~\cite{yang2022vision_tcl} further leverages cross-modal and intra-modal self-supervision simultaneously to obtain multi-modal interactions more easily, leading to the most recent state-of-the-art performance.
\par
Despite impressive performance of ALBEF and TCL, such methods may focus on specific objects of interest from training data, while neglecting other critical objects (see Figure~\ref{fig:teaser}), thus leading to focal bias problem.
This is mainly due to the information disparity between image and text.
For example, as shown in Figure~\ref{fig:teaser}, the corresponding training text of the top row figure is ``\textit{The chair is purple}'', which only describes parts of the image.
The image encoder needs to pay more attention to the text-relevant area to facilitate the alignment between image and text while neglecting some irrelevant regions.
However, for VLP models, without any textual prior, the image feature should be task-agnostic, and all critical objects in an image should be attended to.
To alleviate such bias, in this work, we introduce a generative task of pixel-level reconstruction, relying on a comprehensive understanding of the image. 
\par
Pixel-level image reconstruction with the encoder-decoder paradigm is prevalent in representation learning~\cite{vincent2008extracting, kingma2013auto}.
Recently, with asymmetric encoder-decoder architecture, masked autoencoders (MAE)~\cite{he2022masked} exhibits high generalizability and remarkable performance in vision tasks.
The simple but insightful pre-training strategy, \ie, masking a very high portion of random patches, exploits image redundancy and forces the model to acquire comprehensive understanding of the image.
Inspired by MAE, we adopt the masking-then-predicting paradigm and introduce a pixel-level reconstruction task for alleviating the focal bias problem.
\par
In this paper, we propose a novel VLP framework called Vision-Language Masked AutoEncoder (VLMAE), which is composed of two uni-modal encoders to extract visual and textual features, an image decoder for pixel-level reconstruction, and a fusion learner to aggregate the multi-modal features.
VLMAE employs image-text contrastive learning and visual generative learning simultaneously.
Contrastive learning promotes the alignment of image-text pairs and enables the model to obtain global semantic representations, and generative learning facilitates the model to acquire more comprehensive and fine-grained understanding.
We propose Regional Masked Image Modeling (RMIM) loss to refine the image-text alignment and facilitate fusion of multi-modal features.
To enhance the model's representation ability, we further propose the Image Feature Reconstruction (IFR) task, which requires the model to predict masked patch features based on visible patch inputs.
Meanwhile, the masking design for the image patches in VLMAE reduces pre-training cost significantly.
\par

Our main contributions can be summarized as:
\begin{itemize}
    \item We propose a vision-language masked autoencoder framework (VLMAE), which utilizes generative learning to alleviate the focal bias problem, providing more comprehensive understanding of the images. 
    \item To further refine the image-text alignment and enhance the model's representation ability, we propose Regional Masked Image Modeling (RMIM) and Image Feature Reconstruction (IFR).
    \item Extensive experiments demonstrate the effectiveness and efficiency of VLMAE, which achieves the new state of the art in most downstream vision-language tasks with 20\% reduction of pre-training cost.
\end{itemize}

\section{Related Work}
\subsection{Vison-Language Pre-training}
Most previous vision-language pre-training works can be divided into two categories.
The first category learns image and text embeddings separately with dual uni-modal encoders~\cite{radford2021learning, andonian2022robust}.
Recent methods model the vision-language interaction with contrastive loss on massive noisy paired image-text data
crawled from Internet.
Despite thire simplicity, 
they achieve remarkable performance in cross-modal retrieval tasks with high efficiency.
However, these methods do not perform well in more complex multi-modal downstream tasks, such as visual question answering and visual reasoning, which require the aggregation of visual-textual information and reasoning capabilities. 
They lack the capability to model more complicated
interactions between the two modalities, which is indispensable for multi-modal reasoning.
The second category adopts a transformer-based multi-modal fusion encoder to model the interactions between images and texts~\cite{kim2021vilt, li2021align, zeng2021multi}.
These models achieve superior performance for downstream vision-language reasoning or classification tasks, thanks to the interaction modeling capability of the deep fusion encoder.
Earlier methods rely on a pre-trained object detector for visual region feature extraction, which observably slows down the inference procedure.
Latter methods endeavor to remove the heavy object detector for high inference efficiency.
Among them, SOHO~\cite{huang2021seeing} adopts a vision dictionary to extract compact visual features from the whole image, while ViLT~\cite{kim2021vilt} uses lightweight image and text tokenizers instead of separate uni-modal encoders to achieve faster inference speed.
\par
While these methods of the second category aggregate the multi-modal feature effectively,
they ignore the importance of image-text alignment, hindering the interaction between the visual and textual modalities.
To address this disadvantage, ALBEF~\cite{li2021align} proposes to align the visual and textual representations with contrastive learning before fusing them in the fusion encoder with cross-modal attention. 
Besides, ALBEF adopts pseudo-targets produced by a momentum model to conduct momentum distillation for better noise resistance capability when learning from the noisy web data.
VLMO~\cite{wang2021vlmo} further introduces Mixture-of-Modality-Experts (MOME) Transformer and stage-wise pre-training strategy, leading to better performance in downstream tasks with prominent efficiency.
TCL~\cite{yang2022vision_tcl} also shares the similar aligning-before-fusing spirit of ALBEF but leverages cross-modal and intra-modal self-supervision simultaneously to enhance multi-modal interactions, achieving the most recent state-of-the-art performance.
Though effective, previous vision-language pre-training methods with contrastive learning rely on modeling global semantic interactions and aim to maximize image-text mutual information, resulting in the focal bias problem.
Motivated by recent generative methods in visual self-supervised learning~\cite{he2022masked, chen2022context}, we propose to introduce pixel-level reconstruction task to VLP.
To restore the image, the model needs to pay attention to almost all patches instead of only salient regions.
With such masking-then-predicting paradigm, our model provides more comprehensive understanding of the image and achieves superior performance in downstream multi-modal tasks with
significant pre-training cost reduction.

\subsection{Masked Image Modeling}
Inspired by Masked Language Modeling (MLM)~\cite{devlin2018bert} in NLP, Masked Image Modeling (MIM) is adopted in visual pre-training and has shown impressive results in downstream visual tasks.
Existing MIM works can be classified into two categories. 
Methods in the first category predict discrete tokens generated by VQ-VAE~\cite{van2017neural} or its variants, such as BEiT~\cite{bao2021beit}, mc-BEiT~\cite{li2022mc} and PeCo~\cite{dong2021peco}.
Methods of the second category adopt masking strategies to exploit the redundancy nature of images for visual pre-training.
MaskFeat~\cite{wei2022masked} randomly masks a portion of video sequence and regresses Histograms of Oriented Gradients (HOG) features of masked regions.
MAE~\cite{he2022masked} and SimMIM~\cite{xie2022simmim} predict pixel RGB values directly to promote image pre-training, achieving even better performance than complicatedly designed token classification methods.
\par
Most recently, several concurrent works explore the masked token/patch prediction task for vision-language pre-training.
M3AE~\cite{geng2022multimodal} randomly masks the unified sequence of image patches and text tokens, and encodes the visible image patches into embeddings with the same dimension as the language embeddings to perform joint training of the two modalities. 
Due to the lack of alignment and interaction between the two modalities during pre-training, it is hard to apply directly to various multi-modal downstream tasks.
VLC~\cite{gui2022training} initializes the vision backbone from the pre-trained MAE model, thus avoiding supervised training, and performs intra-modal reconstruction via masked image/language modeling. 
However, the performance in downstream tasks can be sub-optimal due to the lack of elaborate arrangements between MIM and other proxy tasks. 
Meanwhile, VLC suffers from severe pre-training burden due to the large image input size (\eg, 384) and full patches input.
In comparison, our VLMAE adopts a simple yet effective design to simultaneously employ image-text contrastive learning and visual generative learning, achieving SOTA performance with less pre-training cost.

\begin{figure}[t]
\centering
\includegraphics[width=0.95\linewidth, page=1]{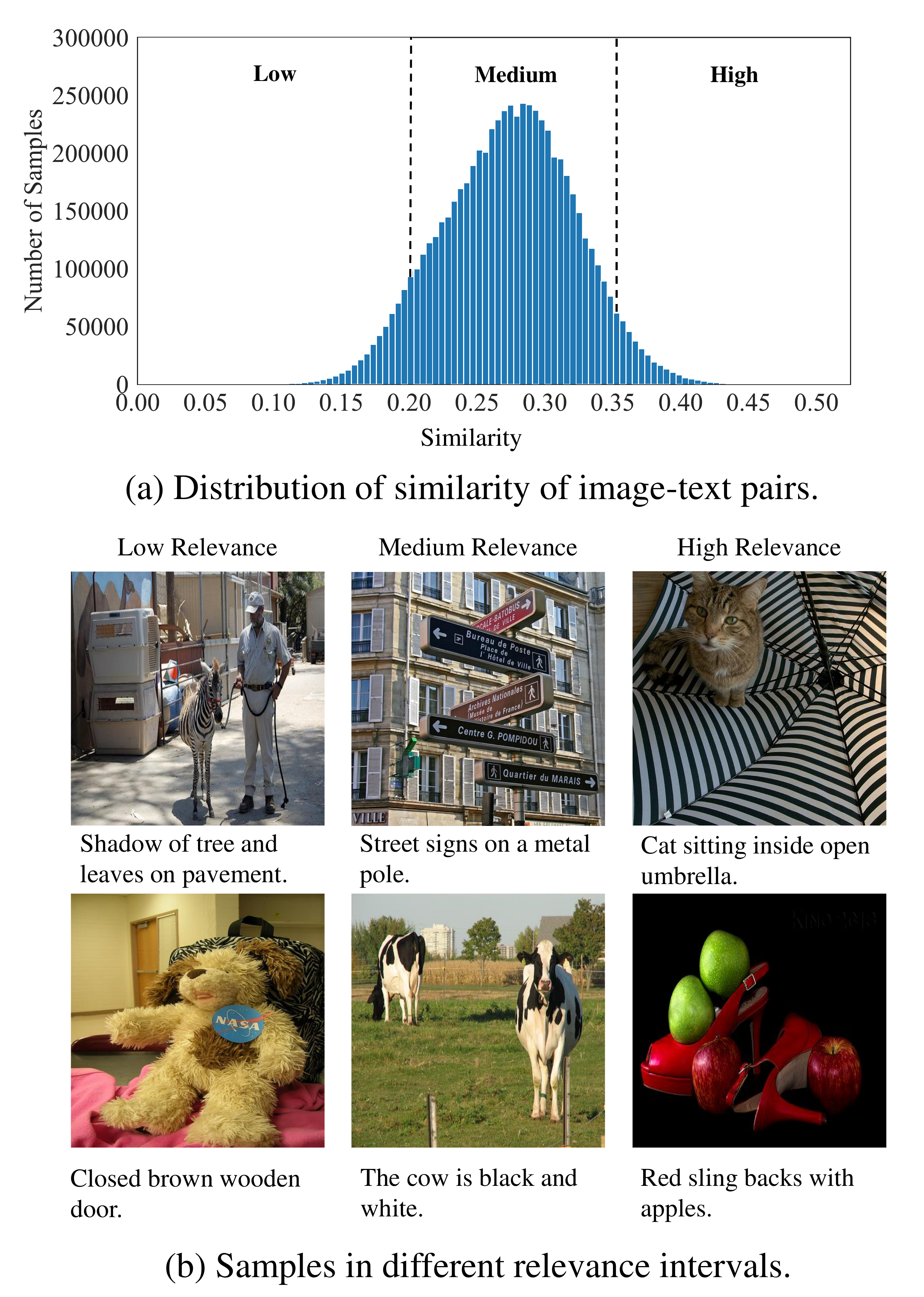}
\caption{Observation on pre-training dataset.}
\label{fig:simi}
\end{figure}

\section{Methodology}
\subsection{Observation}
As image-text dataset plays a fundamental role in vision-language pre-training tasks, we first explore the prevalent pre-training datasets.
Concretely, we calculate the similarity between image and text embedding extracted by a pre-trained CLIP model~\cite{radford2021learning}.
Figure~\ref{fig:simi} shows the distribution of similarity and samples from different intervals. 
In low relevance interval, texts only depict the background or ``unsalient'' regions, and in medium relevance interval, texts describe the ``salient'' objects but ignore some details.
For contrastive learning based methods~\cite{li2021align, yang2022vision_tcl}, this information disparity between image and text may cause the focal bias problem because image encoder tends to pay more attention to the text-relevant area for better alignment between visual and textual embedding. 
Therefore, we introduce generative learning to alleviate such bias during the pre-training phase.

\subsection{Model Architecture}
\begin{figure*}[ht]
  \centering
  \includegraphics[width=0.7\textwidth, page=4]{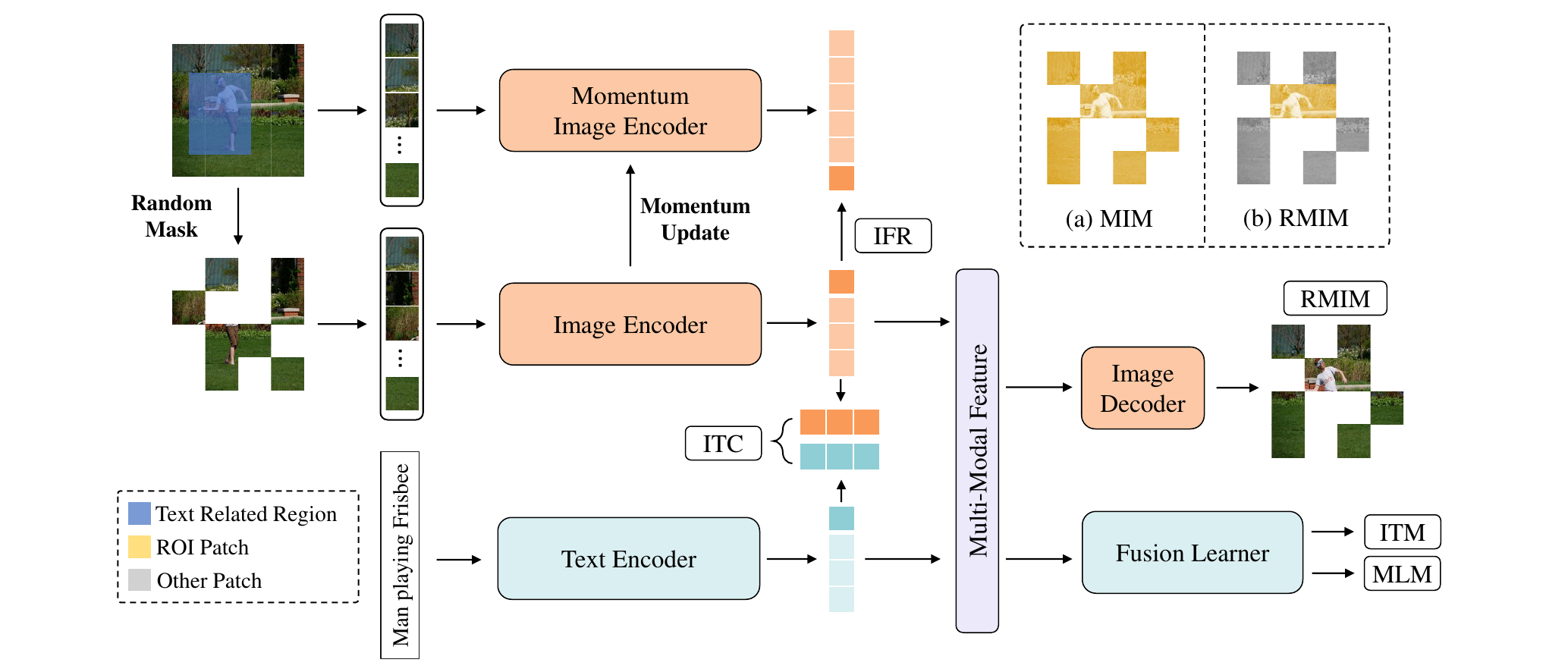}
   \caption{An overview of VLMAE.
   } 
   \label{fig:framework}
\end{figure*}

The overall architecture of our VLMAE model is illustrated in Figure~\ref{fig:framework}. 
VLMAE contains an image encoder $g(\cdot)$ and a text encoder $h(\cdot)$ to extract uni-modal features.
An image decoder is introduced to reconstruct masked patches, and the fusion learner aggregates the vision-language features for multi-modal tasks.
Similar to previous works~\cite{li2021align, yang2022vision_tcl}, for each encoder, we maintain a momentum counterpart $\hat{g}(\cdot)$, $\hat{h}(\cdot)$ and update their parameters following $\theta_{\hat{g}}=m \theta_{\hat{g}}+(1-m) \theta_{g}$.

\subsection{Feature Extraction}
Given an image $I$, we split it into non-overlapping patches and project them into a sequence of patch embeddings, $\left\{x_{i}\right\}_{i=1}^{N}$, where $N$ is the number of patches.
Following MAE~\cite{he2022masked}, we randomly divide the patches into visible patches $\left\{x_{i}^{vis}\right\}_{i=1}^{N - M}$ and invisible patches $\left\{x_{i}^{msk}\right\}_{i=1}^{M}$ according to mask ratio $\alpha$, where $M=\alpha N$. 
During the pre-training phase, we only feed the visible patches into the image encoder, which generates the masked visual representations $\left\{v_{cls}, v_{1}^{vis}, \ldots, v_{N-M}^{vis}\right\}$.
Meanwhile, all patches $\left\{x_{i}\right\}_{i=1}^{N}$ are fed into the momentum image encoder, which generates the visual representations of the whole image $\left\{\hat{v}_{cls}, \hat{v}_{1}, \ldots, \hat{v}_{N}\right\}$. 
For the text input, we feed it into the text encoder and acquire the text representations $\left\{w_{cls}, w_{1}, \ldots, w_{L}\right\}$.
The subscript $cls$ denotes the embedding corresponding to the [CLS] token in both image and text encoders.

\subsection{Regional Mask Image Modeling}
Unlike MAE~\cite{he2022masked}, we aim to reconstruct the invisible patches with both image and text features to facilitate multi-modal information fusion.
Concretely,we concatenate the masked visual representations $\left\{v_{cls}, v_{1}^{vis}, \ldots, v_{N-M}^{vis}\right\}$ and text representations $\left\{w_{cls}, w_{1}, \ldots, w_{L}\right\}$ and feed them into the image decoder. 
Specifically, for the text with bounding box annotation, we propose to apply a regional reconstruction loss to facilitate the aggregation of multi-modal features.
Formally, the reconstruction loss is defined as:
\begin{equation}
\mathcal{L}_\mathrm{rmim}=\mathbb{E}_{(I, T) \sim D} f_{mim}\left(I^{msk} \mid T, I^{vis}\right),
\end{equation}
where $D$ is the pre-training dataset and $T$ is the input text. 
$I^{vis}$ denotes the visible patches and $I^{msk}$ denotes the masked patches in text-relevant region.
For text without bounding box annotation, we treat the whole image as text-relevant region.
The objective $f_{mim}$ calculates the mean squared error (MSE) between the reconstructed and original images at the pixel level.

\subsection{Image Feature Reconstruction}
Because only parts of image patches are fed into the image encoder while the remaining patches are invisible, it causes information insufficiency. 
In order to enhance the representation ability of the model, we propose the image feature reconstruction task, which requires the image encoder to reconstruct the global feature generated by momentum image encoder. 
Concretely, we take the output [CLS] embedding $\hat{v}_{cls}$ of momentum image encoder $\hat{g}(\cdot)$, which is given by the complete image, as the target feature.
The reconstruction loss is denoted as
\begin{equation}
\mathcal{L}_\mathrm{ifr}=\mathbb{E}_{I \sim D} f_{r}(\hat{g}(I)|g(I^{vis})).
\end{equation}
The reconstruction objective $f_{r}$ is L1 loss in practice.

\subsection{Image-Text Contrastive Learning}
To further fuse the vision and language representations, we introduce common vision-language pre-training tasks, namely, image-text contrastive learning (ITC), image-text matching (ITM), and masked language modeling (MLM).
To facilitate the alignment among the features of image and its corresponding texts, we introduce contrastive learning between image feature $v_{cls}$ and text feature $w_{cls}$.
The similarity function is formatted as 
$s(I, T)= \phi_{v}\left(v_{cls}\right)^{\top} \hat{\phi}_{w}\left(\hat{w}_{cls}\right)$ and 
$s(T, I)= \phi_{w}\left(w_{cls}\right)^{\top} \hat{\phi}_{v}\left(\hat{v}_{cls}\right)$, where $\phi(\cdot)$ is a linear projection.
Given a batch of image-text pairs, the image-text and text-image similarities are:
\begin{equation}
\begin{aligned}
\boldsymbol{p}^{\mathrm{i} 2 \mathrm{t}}(I)&=\frac{\exp \left(s\left(I^{vis}, T_{i}\right) / \tau\right)}{\sum_{i=1}^{B} \exp \left(s\left(I^{vis}, T_{i}\right) / \tau\right)}, \\
\boldsymbol{p}^{\mathrm{t} 2 \mathrm{i}}(T)&=\frac{\exp \left(s\left(T, I^{vis}_{i}\right) / \tau\right)}{\sum_{i=1}^{B} \exp \left(s\left(T, I^{vis}_{i}\right) / \tau\right)},
\end{aligned}
\end{equation}
where $B$ is the batchsize and $\tau$ is a learnable temperature parameter.
The image-text contrastive loss is defined as:
\begin{equation}
\mathcal{L}_{\mathrm{itc}}=\mathbb{E}_{(I, T) \sim D}
\left[\mathrm{H}\left(\boldsymbol{y}^\mathrm{i2t}, \boldsymbol{p}^\mathrm{i2t}(I)\right)+\mathrm{H}\left(\boldsymbol{y}^\mathrm{t2i}, \boldsymbol{p}^\mathrm{t2i}(T)\right)\right],
\end{equation}
where $\mathrm{H}(;)$ is the cross-entropy. $\boldsymbol{y}^\mathrm{i2t}$ and $\boldsymbol{y}^\mathrm{t2i}$ are ground truth logits which are guided by momentum distillation following previous work~\cite{li2021align}.

\subsection{Image-Text Matching}
As a widely used training objective in VLP, ITM aims to determine whether a pair of image and text is matched.
Concretely, we take the output [CLS] embedding of fusion learner as the multi-modal representation to predict the correspondence between image and text.
The objective $\mathcal{L}_\mathrm{itm}$ is defined as:
\begin{equation}
\mathcal{L}_\mathrm{itm}=\mathbb{E}_{(I, T) \sim D} \mathrm{H}\left(\boldsymbol{y}^{\mathrm{itm}}, \boldsymbol{p}^{\mathrm{itm}}(I^{vis}, T)\right),
\end{equation}
where $\boldsymbol{p}^{\mathrm{itm}}$ is the predicted probability. $\boldsymbol{y}^{\mathrm{itm}}$ is the grounding truth which evaluates to $1$ iff that the input image and text are matched.

\subsection{Masked Language Modeling}
Following BERT~\cite{devlin2018bert}, MLM aims to predict the masked words based on visual and textual features.
We randomly mask out 15\% tokens in an input sentence. Each token is replaced with the [MASK] token, a random word, or left unchanged, with the probability of 80\%, 10\% and 10\%, respectively.
Denote masked text input as $T^{msk}$, the $\mathcal{L}_\mathrm{mlm}$ is formatted as:
\begin{equation}
\mathcal{L}_{\mathrm{mlm}}=\mathbb{E}_{(I, T) \sim D} \mathrm{H}\left(\boldsymbol{y}^{\mathrm{msk}}, \boldsymbol{p}^{\mathrm{msk}}(I^{vis}, T^{msk})\right),
\end{equation}
where $\boldsymbol{p}^{\mathrm{msk}}$ is the predicted probability of the masked token and $\boldsymbol{y}^{\mathrm{msk}}$ is the ground truth distribution.
\par
The training objective of VLMAE is summarized as follows:
\begin{equation}
\mathcal{L}=\mathcal{L}_\mathrm{rmim}+\mathcal{L}_\mathrm{ifr}+\mathcal{L}_\mathrm{itc}+\mathcal{L}_\mathrm{itm}+\mathcal{L}_\mathrm{mlm}.
\end{equation}

\section{Experiment}
\subsection{Pre-training Datasets}
Following previous work~\cite{chen2020uniter, li2021align, yang2022vision_tcl}, we utilize four common vision and language datasets, MS COCO~\cite{lin2014microsoft}, Visual Genome~\cite{krishna2017visual}, Google Conceptual Captions~\cite{sharma2018conceptual} and SBU Captions~\cite{ordonez2011im2text}, as our pre-training corpus which comprises 4M images and 7M texts. 
We filter the samples in the Visual Genome dataset where the proportion of relevant regions in the images is less than 20\%.
And since the links to some of the images are no longer available in Google Conceptual Captions and SBU Captions, we only access parts of these datasets.
Table~\ref{table:pretraining_data} shows the statistics of the image and text of the pre-training datasets.
\begin{table}
\centering
\footnotesize
\caption{Statistics of pre-training datasets.}
\setlength{\tabcolsep}{6pt}
\begin{tabular}{c|cccc} 
\toprule
& COCO & VG & SBU & CC \\
\midrule
\# images & 113K & 98K & 857K & 2.77M\\
\# text & 567K & 2.85M & 857K & 2.77M\\
\bottomrule
\end{tabular}
\label{table:pretraining_data}
\end{table} 
\begin{table*}[ht]
	\footnotesize
	\setlength\tabcolsep{5pt}
	\begin{center}
		\begin{tabular}{l|c|cccccc|cccccc}
            \toprule 
                \multirow{3}{*}{Method} &
                \multirow{3}{*}{\#Images} & \multicolumn{6}{c}{MSCOCO (5K)} & \multicolumn{6}{c}{Flickr30K (1K)}\\
                
                & & \multicolumn{3}{c}{Text Retrieval} & \multicolumn{3}{c}{Image Retrieval} &
                \multicolumn{3}{c}{Text Retrieval} & \multicolumn{3}{c}{Image Retrieval} \\
            
                & & R@1 & R@5 & R@10 & R@1 & R@5 & R@10
                & R@1 & R@5 & R@10 & R@1 & R@5 & R@10 \\
                \midrule
            
                
                
                OSCAR  & 4M & 70.0 & 91.1 & 95.5 &  54.0 & 80.8 & 88.5 & \xmark  & \xmark & \xmark & \xmark & \xmark & \xmark   \\
                
                ViLT  & 4M & 61.5 & 86.3 & 92.7  & 42.7 & 72.9 & 83.1 & 83.5 & 96.7 & 98.6 &  64.4 & 88.7 & 93.8 \\
                                VLC  & 4M &  71.3 & 91.2 & 95.8 & 50.7 & 78.9 & 88.0 & 89.2 & 99.2 & 99.8 & 72.4 &  93.4 & 96.5 \\

                ALBEF  & 4M &  73.1 & 91.4 & 96.0 & 56.8 & 81.5 & 89.2 & 94.3 & 99.4 & 99.8 & 82.8 &  96.7 & 98.4 \\
                
                 TCL & 4M & 75.6 & 92.8 & 96.7 & 59.0 & 83.2 & 89.9 &  94.9 & 99.5 & 99.8 & \textbf{84.0} & \textbf{96.7} & \textbf{98.5}\\

                Ours & 4M & \textbf{77.3} & \textbf{93.6} & \textbf{97.4} & \textbf{59.6} & \textbf{83.6} & \textbf{90.3} &  
                            \textbf{95.2} & \textbf{99.6} & \textbf{99.9} & 83.6 & 96.6 & \textbf{98.5}\\
                
                \bottomrule
            \end{tabular}
	\end{center}
	\caption{Performance comparison of fine-tuned image-text retrieval on Flickr30K and COCO datasets. 
	}
	\label{table:fine_tune}
\end{table*}

\subsection{Implementation Details}
In VLMAE, we apply a 12-layer ViT-B/16~\cite{dosovitskiy2020image} as our image encoder, which is initialized by ImageNet 1K pre-trained weights.
Following~\cite{he2022masked}, we take a 4-layer transformer with 512 dim as our image decoder. 
The text encoder is implemented by the first 6 layers of $\text{BERT}_{base}$ and the last 6 layers are taken as the fusion learner.
During the pre-training stage, we train the model for 30 epochs with a total batch size of 512 on 8 GPUs. 
We utilize AdamW~\cite{loshchilov2017decoupled} optimizer with weight decay of 0.02.
The learning rate is warmed-up to 1e-4 in the first 1000 iterations and decayed to 1e-5 following a cosine schedule.
Following~\cite{li2021align, yang2022vision_tcl}, we crop the image into $256\times256$ and apply random color jittering, random grayscale conversion, random Gaussian Blur, random horizontal flip, and RandAugment~\cite{cubuk2020randaugment} on it.
During the fine-tuning stage, we increase the image resolution to $384\times384$.

\subsection{Downstream Tasks}
\subsubsection{Image-Text Retrieval}
Image-Text Retrieval consists of two subtasks, namely, image-to-text retrieval (TR) and text-to-image retrieval (IR). 
We evaluate our model on Flickr30K~\cite{plummer2015flickr30k} and COCO~\cite{lin2014microsoft} dataset. 
In testing data, Flickr30K contains 1K images and 5K texts, and COCO contains 5K images and 25K texts.
Following the same protocol as ALBEF~\cite{li2021align}, we conduct experiments on the fine-tuning and zero-shot settings.
In the fine-tuning setting, we fine-tune our pre-trained model on training data and test it on validation/test data.
In the zero-shot setting, following~\cite{li2021align, yang2022vision_tcl}, we fine-tune the model on COCO and test it on Flickr30k.

\subsubsection{Visual Question Answering (VQA)}
In the visual question answering~\cite{goyal2017making} task, given an image and a text question, the model is expected to understand features of both modals and provide a textual answer.
Following the same setting in~\cite{li2021align}, we treat this task as a generation problem and introduce an answer decoder on top of the fusion learner.
Given the multi-model features, the answer decoder is required to generate the answer from 3192 candidates. 

\subsubsection{Visual Entailment (SNLI-VE)}
visual entailment~\cite{xie2019visual} task requires the model to judge the relation between an image and a text is entailment, neutral, or contradictory.
Following the same setting in~\cite{li2021align}, we consider this task as a three-way classification problem and generate the prediction using an MLP on the representation of [CLS] token of the text decoder.
\begin{table}[!t]
\footnotesize
\centering	
\begin{tabular}	{l | c  c  c  c  c  c  }
\toprule	 	
\multirow{2}{*}{Method} &  \multicolumn{3}{c}{TR}& \multicolumn{3}{c}{IR} \\
	 
 & R@1 &R@5&R@10& R@1 &R@5&R@10 \\
\midrule
ALBEF & 90.5 & 98.8 & 99.7 & 76.8 & 93.7 & 96.7 \\
TCL & 93.0 & \textbf{99.1} & 99.6 & \textbf{79.6} & \textbf{95.1} & \textbf{97.4} \\
Ours & \textbf{93.4} & \textbf{99.1} & \textbf{99.8} & 78.6 & 94.5 & 97.3 \\
\bottomrule
\end{tabular}
\caption{Zero-shot image-text retrieval results on Flickr30K.}
\label{tbl:retrieval_zeroshot}
\end{table}		

\subsubsection{Visual Reasoning (NLVR2)}
In visual reasoning~\cite{suhr2018corpus_nlvr} task, given an image-text pair, the model aims to predict whether a text describes a pair of images. 
Following~\cite{li2021align}, we evaluate our model on NLVR2 dataset, which consists of 107,292 image-text examples, and extend the model to suit paired images input.

\subsubsection{Visual Grounding (VG)}
In visual grounding task, the model aims to localize the region in an image corresponding to a specific textual description. 
Following~\cite{li2021align}, we study this task on weakly-supervised setting and evaluate our model on RefCOCO+~\cite{yu2016modeling_grounding} dataset.
Concretely, we fine-tune the pre-trained model with image-text pairs without any bounding box annotations.
During inference, we generate heatmaps by Grad-CAM~\cite{selvaraju2017grad} and rank the proposals generated by MAttNet~\cite{yu2018mattnet} based on these heatmaps. 

\subsection{Evaluation on Image-Text Retrieval}
We first evaluate our model on image-text retrieval task, which is the most common downstream task in vision-language pre-training.
Table~\ref{table:fine_tune} and Table~\ref{tbl:retrieval_zeroshot} report the results on fine-tuned and zero-shot settings, respectively.
In fine-tuned image-text retrieval, our model surpasses the previous state-of-the-art TCL~\cite{yang2022vision_tcl} on MSCOCO dataset, reaching 77.3\% and 59.6\% in terms of TR@1 and IR@1, respectively.
On Flickr30k dataset, VLMAE has a comparable performance with TCL and outperforms previous work ALBEF~\cite{li2021align}.
In the zero-shot setting, VLMAE surpasses TCL on text retrieval subtask but has an inferior performance on image retrieval.
We speculate that this is because, compared with other methods~\cite{li2021align}, the image encoder in VLMAE only sees 50\% patches during the pre-training stage. 
According to the finding in MAE~\cite{he2022masked}, unlike contrastive learning models, generative learning models are not prone to saturation in training, and a longer training schedule can further boost model performance.

\subsection{Evaluation on VQA, VE, and VR}
\begin{table}
\footnotesize
\setlength\tabcolsep{2.8pt}
\begin{center}
\begin{tabular}{l|cc|cc|cc}
\toprule 
\multirow{2}{*}{Method} 
& \multicolumn{2}{c}{VQA} 
& \multicolumn{2}{c}{NLVR$^2$} 
&\multicolumn{2}{c}{SNLI-VE}\\
& test-dev & test-std & dev & test-P & val & test \\
\midrule
OSCAR   & 73.2  & 73.4  & 78.1  & 78.4  &    \xmark   & \xmark \\
ViLT    & 71.3  &  \xmark     & 75.7  & 76.1  &  \xmark     & \xmark \\
VLC     & 74.0  & 74.0  & 77.7  & 79.0  & \xmark  & \xmark  \\
ALBEF   & 74.5  & 74.7  & 80.2  & 80.5  & 80.1  & 80.3  \\
TCL     & 74.9  & 74.9  & \textbf{80.5}  & \textbf{81.3}  & \textbf{80.5}  & \textbf{80.3} \\
Ours    & \textbf{75.3}  & \textbf{75.4}      & \textbf{80.5}  & 81.2  & 80.3  & \textbf{80.3} \\
\bottomrule
\end{tabular}
\end{center}
\caption{Performance comparison on vision question answering, visual reasoning, visual entailment.}
\label{table:vqa}
\end{table}
Table~\ref{table:vqa} shows the result of visual question answering, visual entailment, and visual reasoning tasks which are classical multi-modal tasks and require the model to exploit both image and text information.
VLMAE outperforms TCL~\cite{yang2022vision_tcl} on VQA task and achieves state-of-the-art performance with 75.3\% and 75.4\% scores on test-dev and test-std.
On VE and VR tasks, VLMAE has a comparable performance with TCL, reaching 81.2\% on NLVR test-P and 80.3\% on SNLI-VE test.

\subsection{Evaluation on Visual Grounding}
\begin{table}[!t]
\centering	
\setlength\tabcolsep{4pt}
\begin{tabular}	{l  |  c c c }
\toprule
Method & Val & TestA & TestB \\
\midrule
ALBEF   & 58.5  & 65.9  & 46.3 \\		
Ours    & \textbf{62.3}  & \textbf{71.6}  & \textbf{50.7} \\
\bottomrule
\end{tabular}
\caption{Visual grounding on RefCOCO+ dataset.}
\label{tbl:grounding}	
\end{table}
                
                

\begin{table}[t]
	\footnotesize
	\begin{center}
		\begin{tabular}{l|cccc}
            \toprule 
\multirow{2}{*}{Module} & \multicolumn{2}{c}{MSCOCO} & \multicolumn{2}{c}{Flickr30K} \\
& TR & IR & TR & IR \\
\midrule
Baseline &87.9 &76.6 &97.9 &91.8   \\
+ MIM&88.9 &77.3 &98.2 &92.1   \\
+ RMIM&89.0 &77.5 &98.1 &92.6   \\
+ RMIM + IFR&\textbf{89.4} &\textbf{77.8} &\textbf{98.2} &\textbf{92.9}   \\
\bottomrule
\end{tabular}
\end{center}
\caption{Ablation study of Modules. Results are reported in terms of  the average score of TR and IR.}
\label{table:ablation_pool}
\end{table}

\begin{figure}[t]
  \centering
  \includegraphics[width=0.8\linewidth, page=1]{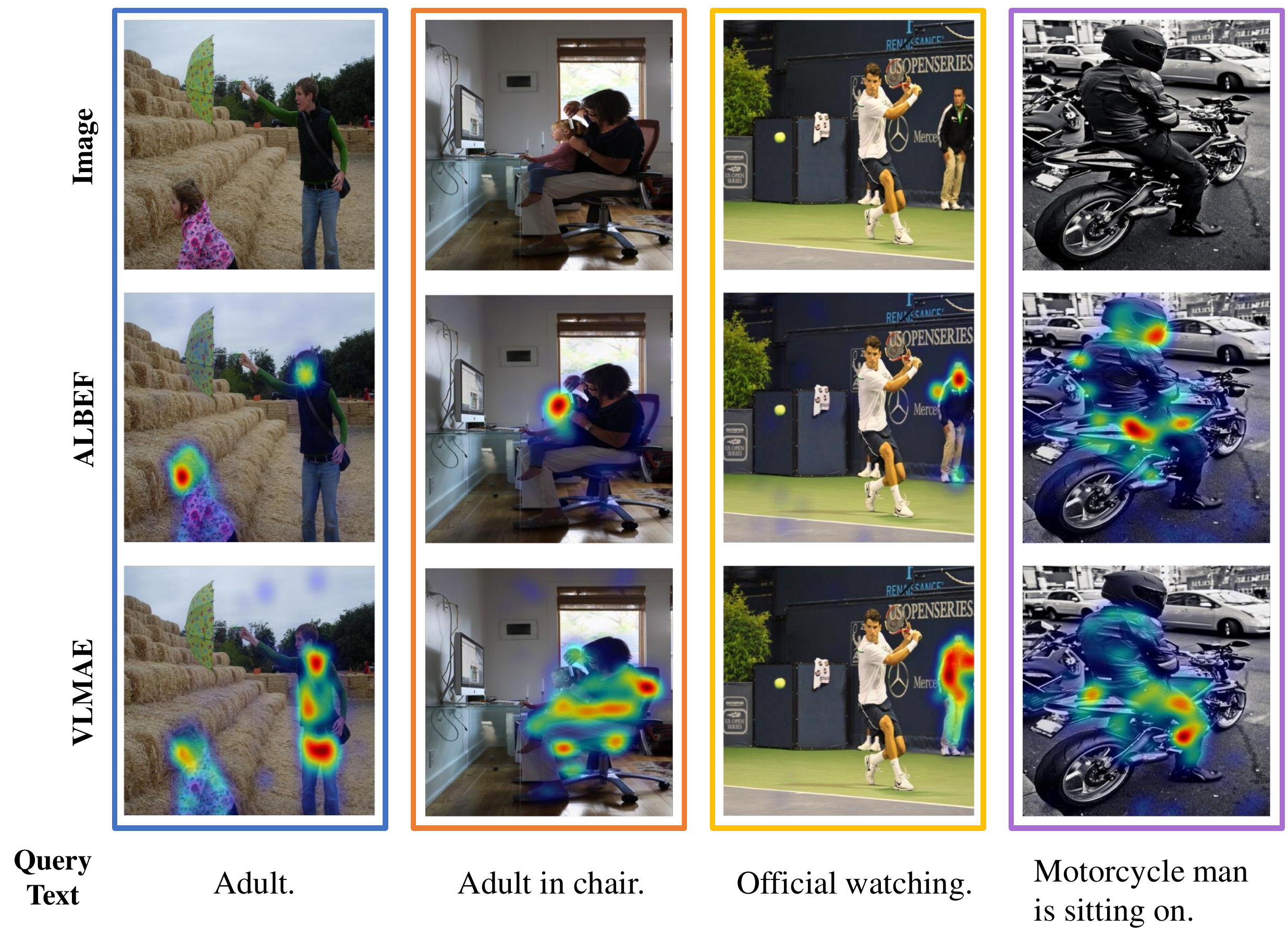}
   \caption{Grad-CAM visualization in visual grounding task.} 
   \label{fig:grounding}
\end{figure}

As shown in Table~\ref{tbl:grounding}, VLMAE surpasses the previous state-of-the-art ALBEF~\cite{li2021align} by a large margin, reaching 71.6\% and 50.7\% accuracy on Test data. 
Similar to the finding in CAE~\cite{chen2022context}, we argue that the reconstruction of pixels in an image makes the model focus on almost every patch instead of only the salient region, resulting in better performance on location tasks (see Section~\ref{sec:quali_assess} and Figure~\ref{fig:grounding}).

\subsection{Ablation Study}
In this section, we conduct ablation studies on image-text retrieval task to validate the effectiveness of the newly proposed RMIM and IFR.
Table~\ref{table:ablation_pool} shows the results on both MSCOCO and Flickr30k datasets.
We consider the VLMAE trained with only ITC, ITM, and MLM loss as our baseline model.
With the addition of MIM loss, model achieves better performance on both IR and TR tasks.
We argue that the reconstruction of pixels can facilitate the model to capture more comprehensive feature of the image. 
Moreover, introducing regional MIM loss further improves the model performance, which can be attributed to better image-text alignment.
With the reconstruction of the image feature with only visible patches input, our model has a stronger representation ability, resulting in SOTA performance on image-text retrieval tasks.
\begin{table}
	\footnotesize
	\begin{center}
		\begin{tabular}{c|cccc}
            \toprule 
                \multirow{3}{*}{Mask Ratio} &
                \multicolumn{4}{c}{Fine-Tune}\\
                 & \multicolumn{2}{c}{MSCOCO} & \multicolumn{2}{c}{Flickr30K} \\
                
                & TR & IR & TR & IR \\
                \midrule
                25\% &89.1  &\textbf{78.0}  &\textbf{98.4}  &92.6  \\
                50\% &\textbf{89.4}  &77.8  &98.2  &\textbf{92.9}  \\
                75\% &88.1  &76.7  &98.0  &91.6  \\

                \bottomrule
            \end{tabular}
	\end{center}
	\caption{Exploration of mask ratio. Results are reported in terms of the average score of TR and IR.}
	\label{table:mask_ratio}
\end{table}

\begin{figure}[t]
\centering
\includegraphics[width=0.85\linewidth, page=5]{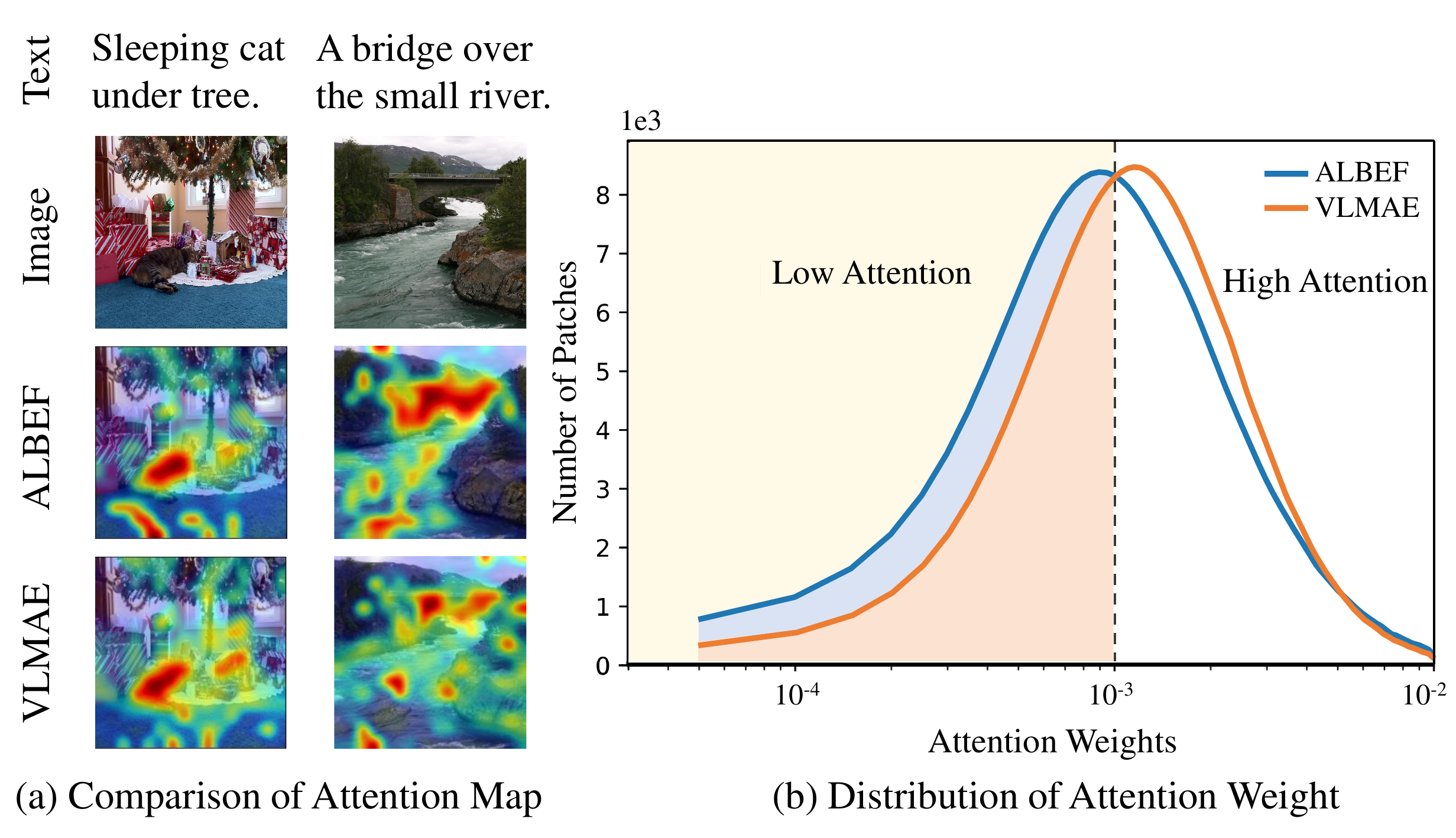}
\caption{Comparison of attention. 
}
\label{fig:attn}
\end{figure}
We also explore the effect of the mask ratio, which is a trade-off between training efficiency and model performance.
Table~\ref{table:mask_ratio} shows the results with different mask ratios.
Compared with ImageNet~\cite{russakovsky2015imagenet} dataset, the content of images in vision-language pre-training datasets is more complicated.
We argue that a too large mask ratio (\eg, 75\%) may cause severe information loss, leading to a performance drop.
In practice, we choose a 50\% mask ratio to balance training efficiency and performance.
\begin{table}[t]
\footnotesize
\begin{center}
\begin{tabular}{c|cccccc}
\toprule 
\multirow{2}{*}{Method} 
& \multirow{2}{*}{MACs}      
& Epoch
& \multicolumn{2}{c}{MSCOCO} 
& \multicolumn{2}{c}{Flickr30K} \\
&&Time& TR & IR & TR & IR \\
                \midrule
ALBEF & 60.5G & 7.7h    &86.8  &75.8  &97.8  &92.6  \\
TCL   & 60.6G & 7.8h    &88.4  &77.4  &98.1  &\textbf{93.1}  \\
VLMAE & 49.6G & 6.2h    &\textbf{89.4}  &\textbf{77.8}  &\textbf{98.2}  &92.9  \\
                \bottomrule
            \end{tabular}
	\end{center}
	\caption{Comparison of training consumption.}
	\label{table:time}
\end{table}

\subsection{Qualitative Assessment}
\label{sec:quali_assess}
To demonstrate the effectiveness of our model, following ALBEF~\cite{li2021align}, we visualize the cross-attention maps in visual grounding task.
As shown in Figure~\ref{fig:grounding}, compared with ALBEF, VLMAE can identify the relevant region more precisely and pay attention to the whole relevant area.

\subsection{Investigation About Focal Bias}
To demonstrate that VLMAE can alleviate the focal bias problem, we explore the attention weights between the [CLS] token and the patch tokens of the image encoder.
Results are shown in Figure~\ref{fig:attn}.
Since the text in pre-training dataset only depicts parts of the image (\eg, ``A bridge over the small river''), ALBEF neglects other irrelevant regions (\eg, ``River bank'').
However, thanks to generative learning, VLMAE pays attention to most patches and provides more comprehensive understanding, which is fundamental to downstream tasks. 
Moreover, according to the statistic on attention weights, VLMAE has fewer patches in the low-attention interval, which means less information is neglected, and extracted features are more unbiased.

\subsection{Pre-training Efficiency}
To demonstrate the efficiency of our proposed method, in Table~\ref{table:time}, we compare VLMAE with previous works~\cite{li2021align, yang2022vision_tcl} in terms of MACs, time of epoch, and image-text retrieval result.
The MACs of each model in one forward pass (divided by batchsize 64) is calculated by the thop package\footnote{https://github.com/Lyken17/pytorch-OpCounter}.
Time spent per epoch is tested on 8 NVIDIA V100 GPUs.
As only half of the image patches are fed into the image encoder during the pre-training stage, VLMAE achieves state-of-the-art performance with lower memory consumption and shorter pre-training time.

\section{Conclusion}
In this paper, we propose a vision-language masked autoencoder framework (VLMAE).
VLMAE utilizes an asymmetric encoder-decoder architecture, consisting of a fusion learner for aggregating multi-modal information and a lightweight image decoder for reconstructing pixels.
Due to its simple yet effective design, VLMAE can successfully handle the focal bias problem in pre-training, providing a more comprehensive understanding of the image.
Moreover, VLMAE utilizes Regional Masked Image Modeling to stabilize pixel-level reconstruction and Image Feature Reconstruction to improve the model's representation ability.
Extensive experiments on various downstream vision-language benchmarks demonstrate that VLMAE surpasses existing state-of-the-art methods while prominently alleviating the training cost in the pre-training stage.

\bibliography{aaai23}
\clearpage
\appendix
\vspace{8pt}
\begin{center} \begin{Large}{\textbf{Supplementary Material}} \end{Large} \end{center}
\vspace{4pt}

\section{Model Architecture}
VLMAE consists of two uni-modal encoders to extract visual and textual features, an image decoder for the image reconstruction, and a fusion learner for multi-modal aggregation.
Concretely, we build the text encoder and fusion learner based on standard $\text{BERT}_{base}$~\cite{devlin2018bert}.
For image encoder, we employ a standard ViT-B/16~\cite{dosovitskiy2020image} with random masking strategy~\cite{he2022masked}.
Following MAE~\cite{he2022masked}, we adopt a lightweight vision transformer as our image decoder.
Details about the architecture are illustrated in Table~\ref{tab:supp_arch}.

\section{Discussion about the attention map}
\subsubsection{Desirable Features for Vision-Language tasks.}
Different from single-label classification tasks, where the model is only required to pay attention to the most salient object in an image, vision-language tasks are more challenging.
The first challenge is that vision-language tasks are typically open-set.
Without text or question prior, the model cannot identify the region of interest because anything in an image could potentially be the hint to answer.
Moreover, the complexity of contents in most real-world images requires the model to provide comprehensive understanding (\eg, objects and their relations) of an image.
Therefore, desirable features for vision-language tasks should incorporate information about objects as much as possible, which means the model should pay attention to most critical objects in an image.

\subsubsection{Why Not Use Grad-CAM?}
Compared with Grad-CAM, attention map is task-agnostic and more general for various tasks. 
To acquire Grad-CAM of an image, we have to calculate the gradient according to a specific loss function, which means Grad-CAM is task-specific.
Moreover, for a pre-training method, especially self-supervised methods, pre-text tasks are usually meaningless in practice.
In contrast, attention maps are easy to acquire without any task prior and reflect the model's attention directly, which is more suitable for exploring the representation ability of image encoder.
\subsubsection{More Visualizations.}
\begin{figure*}[h]
  \centering
  \includegraphics[width=1\linewidth, page=1]{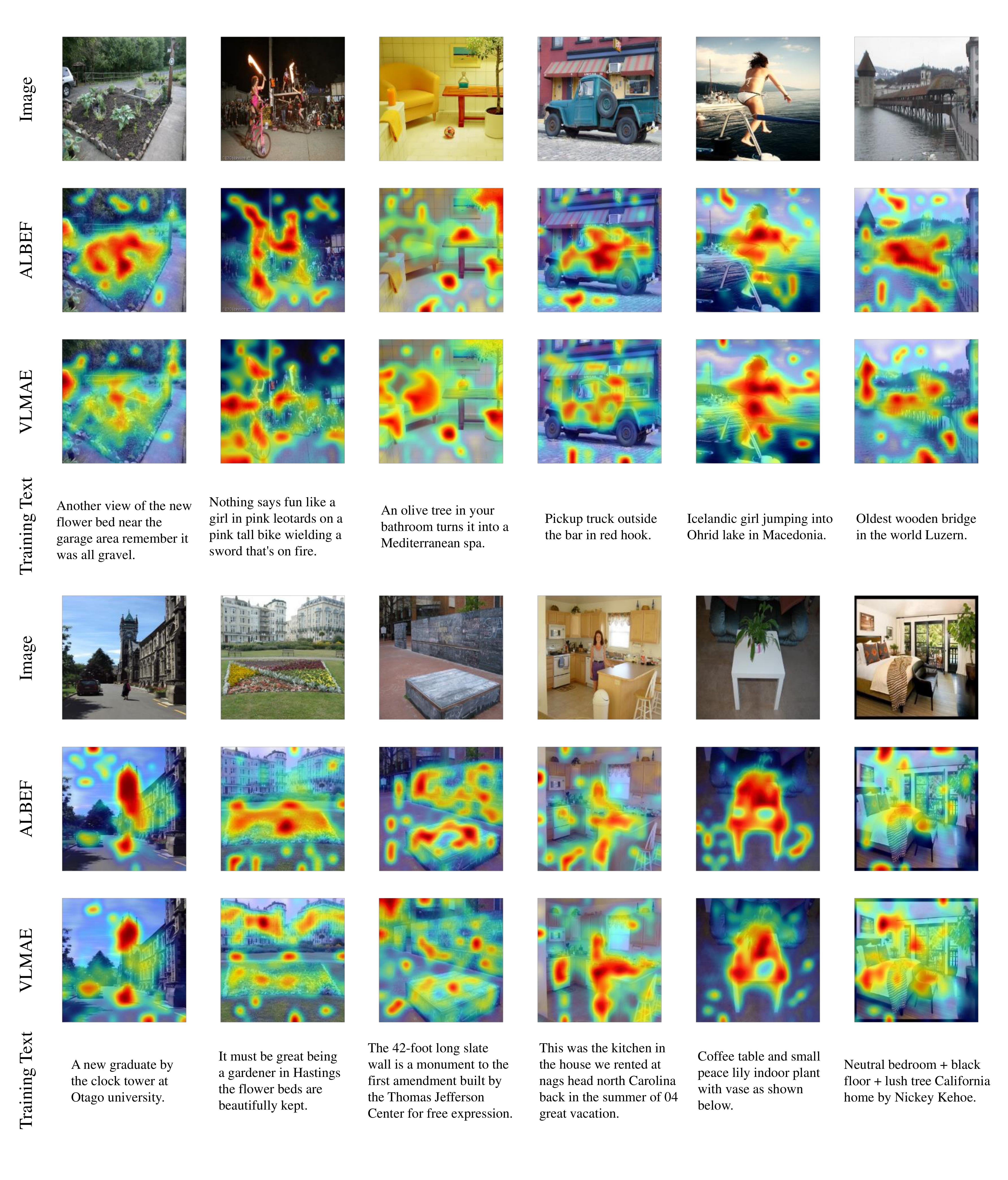}
  \caption{Additional attention maps of pre-training samples.} 
  \label{fig:supp_attn}
\end{figure*}

In Figure~\ref{fig:supp_attn}, we visualize attention maps of more pre-training samples to demonstrate that VLMAE can alleviate the focal bias problem and provide more comprehensive understanding.
In contrast to ALBEF, which mainly focuses on the training-text relevant areas, VLMAE attends to almost all critical objects in an image. 


\begin{table}[t]
\tablestyle{6pt}{1.02}
\footnotesize
\begin{tabular}{l|l|l}
module & config & value \\
\shline
\multirow{7}{*}{Image Encoder}&patch size & 16 \\ 
&hidden layer & 12 \\
&attention heads & 12 \\
&hidden size & 768 \\
&mlp ratio & 4 \\
&activation function & GELU \\
&norm & LayerNorm \\ 
\hline
\multirow{6}{*}{Text Encoder}&hidden layer & 6 \\
&attention heads & 12 \\
&hidden size & 768 \\
&mlp ratio & 4 \\
&activation function & GELU \\
&norm & LayerNorm \\ 
\hline
\multirow{6}{*}{Image Decoder}&hidden layer & 4 \\
&attention heads & 8 \\
&hidden size & 512 \\
&mlp ratio & 4 \\
&activation function & GELU \\
&norm & LayerNorm \\ 
\hline
\multirow{6}{*}{Fusion Learner}&hidden layer & 6 \\
&attention heads & 12 \\
&hidden size & 768 \\
&mlp ratio & 4 \\
&activation function & GELU \\
&norm & LayerNorm \\ 
\shline
\end{tabular}
\caption{\textbf{Model Architecture.}}
\label{tab:supp_arch}
\end{table}

\begin{figure*}[h]
  \centering
  \includegraphics[width=1\linewidth, page=2]{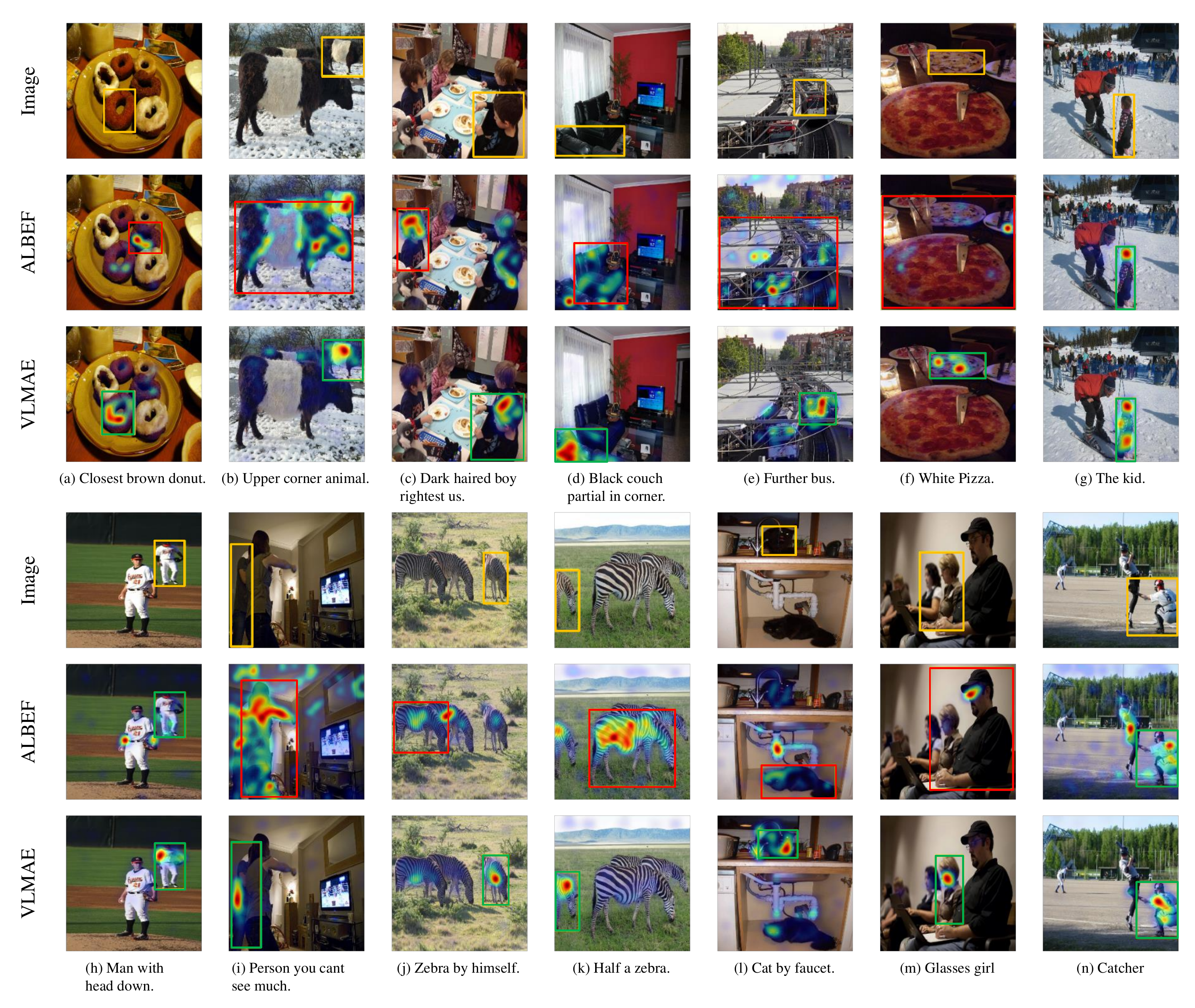}
  \caption{Additional results in visual grounding task.
  Orange: ground truths. Green: correct predictions. Red: incorrect predictions.} 
  \label{fig:supp_vg}
\end{figure*}
\section{Visual Grounding Results}
Thanks to generative learning, VLMAE can recognize most critical objects in an image and provide more comprehensive features, facilitating downstream tasks like visual grounding.
To further demonstrate the superior performance of VLMAE on localization tasks, in Figure~\ref{fig:supp_vg}, we provide more results in visual grounding task. 
Compared with ALBEF~\cite{li2021align}, VLMAE is capable of better identifying relative positions (\eg, (a), (b), (c), (d) and (e)), recognizing the object properties (\eg, (f) and (m)), detecting whole relevant areas (\eg, (g)), distinguishing similar objects (\eg, (i), (j) and (k)), determining the subject of descriptions (\eg, (m)) and suppressing the noise regions (\eg, (g), (h), and (n)). 
\end{document}